\documentclass[sn-mathphys-num]{sn-jnl}

\usepackage{graphicx}%
\usepackage{multirow}%
\usepackage{amsmath,amssymb,amsfonts}%
\usepackage{amsthm}%
\usepackage{mathrsfs}%
\usepackage[title]{appendix}%
\usepackage{xcolor}%
\usepackage{textcomp}%
\usepackage{manyfoot}%
\usepackage{booktabs}%
\usepackage{algorithm}%
\usepackage{algorithmicx}%
\usepackage{algpseudocode}%
\usepackage{listings}%

\theoremstyle{thmstyleone}%
\theoremstyle{thmstyletwo}%

\theoremstyle{thmstylethree}%

\begin{document}

\title[SSL AD Biomarkers]{A Cautionary Tale of Self-Supervised Learning for Imaging Biomarkers: Alzheimer's Disease Case Study}


\author*[1]{\fnm{Maxwell} \sur{Reynolds}}\email{mar398@pitt.edu}

\author[2]{\fnm{Chaitanya} \sur{Srinivasan}}

\author[2]{\fnm{Vijay} \sur{Cherupally}}

\author[2]{\fnm{Michael} \sur{Leone}}

\author[1]{\fnm{Ke} \sur{Yu}}

\author[3]{\fnm{Li} \sur{Sun}}

\author[1]{\fnm{Tigmanshu} \sur{Chaudhary}}

\author[2]{\fnm{Andreas} \sur{Pfenning}}

\author[3]{\fnm{Kayhan} \sur{Batmanghelich}}

\affil*[1]{\orgdiv{Department of Biomedical Informatics}, \orgname{University of Pittsburgh}, \orgaddress{\street{5607 Baum Blvd.}, \city{Pittsburgh}, \postcode{15206}, \state{PA}, \country{United States}}}

\affil[2]{\orgdiv{Department of Computational Biology}, \orgname{Carnegie Mellon University}, \orgaddress{\street{5000 Forbes Ave,}, \city{Pittsburgh}, \postcode{15213}, \state{PA}, \country{United States}}}

\affil[3]{\orgdiv{Department of Electrical and Computer Engineering}, \orgname{Boston University}, \orgaddress{\street{8 St. Mary's St.}, \city{Boston}, \postcode{02215}, \state{MA}, \country{United States}}}


\abstract{

Discovery of sensitive and biologically grounded biomarkers is essential for early detection and monitoring of Alzheimer’s disease (AD). Structural MRI is widely available but typically relies on hand-crafted features such as cortical thickness or volume. We ask whether self-supervised learning (SSL) can uncover more powerful biomarkers from the same data. Existing SSL methods underperform FreeSurfer-derived features in disease classification, conversion prediction, and amyloid status prediction. We introduce Residual Noise Contrastive Estimation (R-NCE), a new SSL framework that integrates auxiliary FreeSurfer features while maximizing additional augmentation-invariant information. R-NCE outperforms traditional features and existing SSL methods across multiple benchmarks, including AD conversion prediction. To assess biological relevance, we derive Brain Age Gap (BAG) measures and perform genome-wide association studies. R-NCE-BAG shows high heritability and associations with MAPT and IRAG1, with enrichment in astrocytes and oligodendrocytes, indicating sensitivity to neurodegenerative and cerebrovascular processes.}




\maketitle

\section{Introduction}\label{sec1}
Alzheimer's disease (AD) neuropathology is a process that begins years, even decades, before the onset of clinical symptoms such as confusion and memory impairment \cite{Lloret2019WhenBiomarkers}. Identifying individuals at risk for developing AD is critical when conducting clinical trials for potential treatments \cite{Maheux2023ForecastingDisease}. AD biomarker development has accelerated across several modalities, facilitating AD progression monitoring \cite{Koval2021ADProgression}. Moreover, biomarkers (e.g. from imaging) can serve as \emph{endophenotypes}, or intermediary signals between genetics and disease diagnosis that facilitate a more fine-grained, highly-powered study of disease etiology \cite{Batmanghelich2017ProbabilisticDiagnosis,Xu2017Imaging-wideGWAS}.

A biomarker should be highly sensitive and specific, especially at distinguishing between healthy and prodromal AD patients.
Cognitive tests are less ideal for early detection, as AD may be irreversible by the time symptoms appear \cite{vanderFlier2023TowardsDementia}. While PET and CSF markers for amyloid and tau are sensitive, they are also invasive and/or expensive.
Magnetic resonance imaging (MRI) offers a more cost-effective option than PET and is less invasive than PET and CSF monitoring. However, MRI is generally less sensitive than PET and CSF markers \cite{Frisoni2010TheDisease}. Other markers such as blood-based biomarkers suffer from worse sensitivity than MRI \cite{Nallapu2024PlasmaImpairment}.

\begin{figure*}[!h]
\centerline{\includegraphics[width=\textwidth]{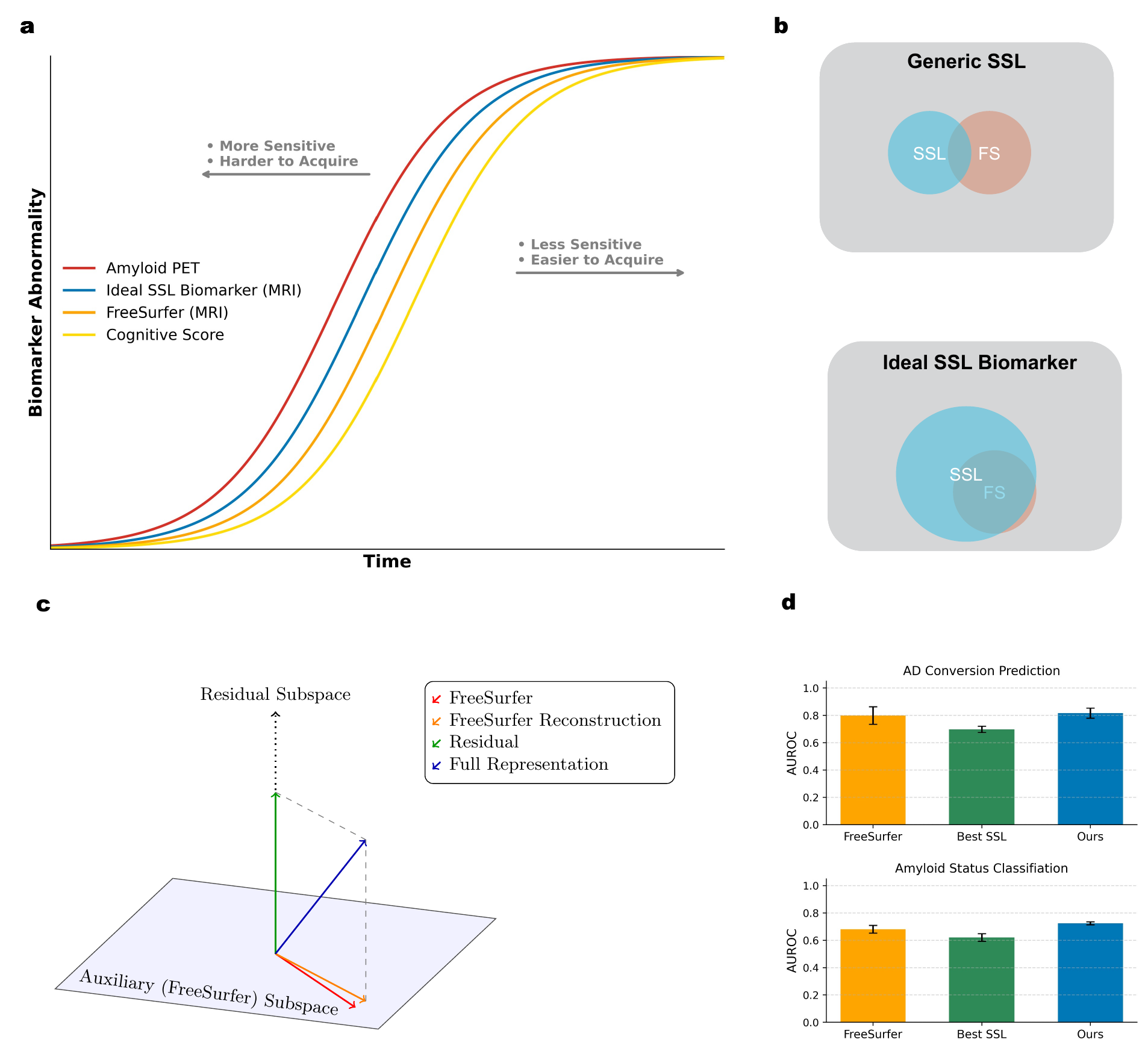}
} 
\caption[Ideal SSL Biomarker]{Ideal SSL Biomarker. a) Alzheimer's Disease (AD) progression curves \cite{Jack2013TrackingBiomarkers} showing the sensitivity of the biomarkers to characterize disease progression. The x-axis shows relative time of biomarker measurements; the y-axis shows the degree of biomarker abnormality. Ideally, MRI-based SSL biomarkers should detect decline earlier than classical imaging biomarkers (e.g., FreeSurfer). Unlike different modalities (e.g. PET), this increased sensitivity would not come at an additional acquisition cost. b) Biomarker information content (blue circle) with different training strategies. Generic SSL (top) ignores FreeSurfer (FS) and therefore does not learn all of the FS information. An ideal SSL biomarker (bottom) learns all FS information and additional relevant information. c) Proposed Method. Auxiliary (FreeSurfer) features are learned (and reconstructed). Next, the information in the residual is maximized via contrastive learning The full representation contains both auxiliary feature information and additional information not captured by auxiliary features. d) AD and amyloid classification performance. The top baseline SSL method significantly underperforms FreeSurfer, while our approach improves on FreeSurfer, indicating a more sensitive AD biomarker.
}
\label{Figure: Ideal Biomarker}
\end{figure*}

An ideal biomarker might have the accessibility of MRI with sensitivity approaching that of PET and CSF markers (see Figure \ref{Figure: Ideal Biomarker}a). Traditionally, MRI biomarkers for AD are scalar structural descriptors such as regional grey matter thickness and volume. Alternatively, image representations can be learned using deep learning.
Supervised learning requires vast labeled datasets and often generalizes poorly \cite{Tendle2021ARepresentations}. In contrast, self-supervised learning (SSL) learns generalizable representations without labels, and has gained traction in medical imaging segmentation \cite{Ren2022LocalAnalysis,Sun2023Self-supervisedAnalysis}, registration \cite{An2022Self-SupervisedImages,Liu2021SAME:Embeddings}, and classification \cite{Zhong2021AImage,Li2021Dual-streamLearning}.
SSL has also been suggested as a helpful framework to create biomarkers for AD and aging \cite{Zhao2021LongitudinalLearning,Kwak2023Self-SupervisedAmyloid-PET,Kim2023LearningImages}.

While SSL presents a promising approach for neuroimaging biomarker discovery \cite{DongRegionalMRI,Kwak2023Self-SupervisedAmyloid-PET,Kim2023LearningImages,Zhao2021LongitudinalLearning,Dadsetan2023RobustLearning}, it surprisingly has not yet been compared directly against traditional structural imaging features. This comparison is critical to establish any clinical utility of SSL. 
We develop evaluation metrics to compare traditional FreeSurfer (FS) cortical thickness and subcortical volume features \cite{Fischl2012FreeSurfer} against four SSL methods.
We find that all SSL methods underperform FS features on our evaluations, raising questions of the utility of generic SSL for AD biomarkers.
 
Based on our initial analysis, we believe FS features contain more useful information about brain morphology which is not entirely captured by any generic SSL training process. An SSL biomarker, on the other hand, should contain more useful information than traditional biomarkers (see Figure \ref{Figure: Ideal Biomarker}b), but this is difficult to ensure with existing SSL objectives. To this end, we develop a new SSL framework which incorporates auxiliary information, such as FS features, into the training procedure.
This method learns the auxiliary information while maximizing residual information content.
We find this method yields more sensitive biomarkers compared to FS and existing SSL approaches.

One critical use of brain imaging biomarkers is to calculate ``Brain Age". The brain age, or age predicted by a model trained on a normative population, indicates a subject's deviation from a typical aging trajectory \cite{J.2023TheAge}. The Brain Age Gap (BAG) is the difference between a subject's predicted brain age and their true age. BAG is a useful metric for overall brain health. It can be trained from a variety of modalities, requires only an age label (i.e. no disease labels), and it is a single scalar value. This allows for straightforward statistical association studies of brain health with genetic markers \cite{Lin2020OlderIllness,J.2023TheAge} and other outcomes \cite{Lin2020OlderIllness,Smith2019EstimationImaging,Cole2018BrainMortality}. Using FS features, representations from a generic SSL method, and representations from our newly proposed method, we create three BAG markers. We compare the three markers with respect to their sensitivity of ground truth neuropathology. We also use the three markers to study the genetic basis for neurodegeneration in older adults and compare findings from each method.

\section{Results}\label{results}
\subsection{High-dimensional Biomarker Sensitivity} \label{Results: Sensitivity}

We first ask whether generic self-supervised representations trained without any domain-specific guidance can serve as sensitive imaging biomarkers. To test this, we evaluate several ``generic'' SSL frameworks developed for natural and medical images and compare their ability to capture biological variation against traditional FreeSurfer (FS) features.
The generic SSL methods are SimCLR \cite{Chen2020ARepresentations}, Barlow Twins \cite{Zbontar2021BarlowReduction}, Models Genesis \cite{Zhou2021ModelsGenesis}, and Longitudinal Self-Supervised Learning (LSSL) \cite{Zhao2021LongitudinalLearning}. SimCLR and Barlow Twins are two contrastive (sample-contrastive and dimension-contrastive, respectively) non-generative methods. Models Genesis and LSSL are reconstruction-based SSL methods specifically designed for medical images. For FS features we use cortical thickness and subcortical volume features acquired using FS recon-all \cite{Fischl2012FreeSurfer}.

We perform a variety of evaluations to test the sensitivity of SSL-based biomarkers compared to traditional cortical thickness and subcortical volume features. The evaluations test a high-dimensional biomarker’s ability to represent relevant biological information related to Alzheimer’s disease status and aging. All evaluations are performed on 2,431 patients from the Alzheimer’s Disease Neuroimaging Initiative (ADNI) \cite{Petersen2010AlzheimersADNI}. We use frozen SSL representations (without supervised fine-tuning to the specific task of interest) and FS features as input to linear classification or regression models specific to each task. Next, we use bootstrap resampling and permutation tests to compare biomarker classification performance. 

\begin{figure*}[!h]
\centerline{\includegraphics[width=\textwidth]{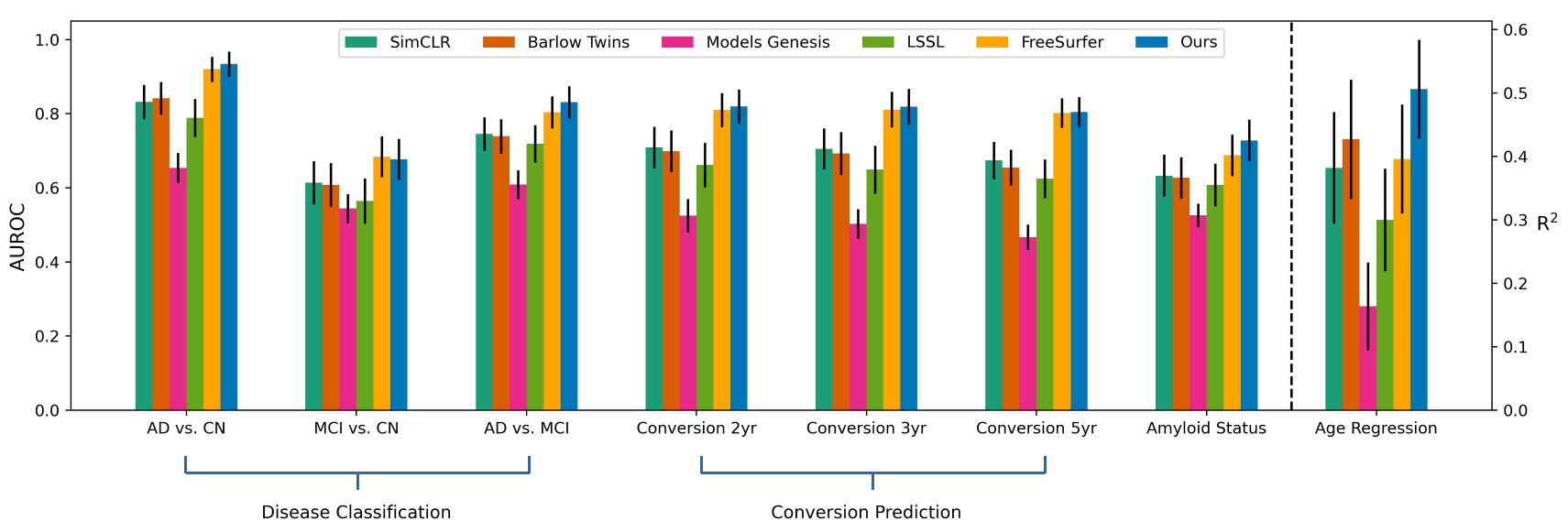}
} 
\caption[Biomarkers Performance Results]{Biomarkers Performance Results. Evaluations showing the sensitivity of various imaging biomarkers are shown for a variety of tasks including disease classification, AD conversion prediction, amyloid status classification, and age regression. Error bars are computed using the standard deviation of 1000 bootstrap samples. 
}
\label{Figure: Biomarker Performance}
\end{figure*}

The first evaluation tests whether different biomarkers are sensitive to a subject's age. Age prediction is a commonly used metric in imaging marker evaluation and can also be used as a proxy for accelerated neurodegeneration in AD and other diseases. We test the age-preditictive capacity of different biomarkers using cognitively normal (CN) subjects from the ADNI dataset. Next, we evaluate the ability of the biomarkers to distinguish subjects based on disease status, framing this as three binary classification tasks: (1) AD vs. CN, (2) MCI vs. CN, and (3) AD vs. MCI. We also test whether biomarkers can be used to predict a future diagnosis of AD. We test conversion prediction performance for three time-scales: 2-year, 3-year, and 5-year. Finally, we examine whether the imaging biomarkers can be used as a proxy for more expensive and sensitive markers (i.e. amyloid). Brain amyloid levels, while not directly visible in a structural image, should be captured by subtle anatomical changes correlated with amyloid accumulation. For this experiment, we test whether representations can classify subjects as being amyloid positive or negative. In all tasks except age prediction, FS outperforms all generic SSL methods $(p < 0.0001)$ (Figure \ref{Figure: Biomarker Performance}).

\subsection{Residual Noise Contrastive Estimation}
Across multiple evaluations, generic SSL representations capture some age and disease-related variation but fall short of the sensitivity of FS-derived features. 
This suggests generic SSL lacks the biological specificity of FS features. To address this, we introduce a new SSL framework designed to incorporate domain-specific information. The key idea is to guide representation learning using auxiliary FS-derived features while still allowing the model to discover additional structure beyond them.
In this approach, representations are learned using two objective functions. First, the representations are trained to be predictive of auxiliary (i.e. FS) features. Second, additional information beyond the auxiliary features is learned via instance-discriminative contrastive learning performed in a residual space that is orthogonal to the auxiliary feature space. We call this method Residual Noise Contrastive Estimation (R-NCE) (Figure \ref{Figure: Ideal Biomarker}c). We use a patch-based convolutional neural network (CNN) to encode structural images and apply this novel loss formulation on both a patch and aggregate image level.

This hybrid strategy closes the performance gap between generic SSL and FS, and in most cases surpasses both (Figure \ref{Figure: Biomarker Performance}). R-NCE improves on all generic SSL methods and FreeSurfer in most biomarker sensitivity metrics. R-NCE is the most sensitive to age-related information, significantly outperforming FS and baseline SSL approaches $(p < .0001)$. Moreover, R-NCE further improves upon FS features in AD vs. CN $(p < 0.0001)$ and AD vs. MCI $(p < 0.0001)$ classification, but slightly underperforms FS in MCI vs. CN $(p = 0.003)$ classification. R-NCE further improves performance in 2-year $(p < 0.0001)$ and 3-year $(p < 0.0001)$ conversion prediction but does not show a statistically significant difference in 5-year conversion prediction compared to FS. Finally, R-NCE shows an improvement in amyloid sensitivity compared to FS features ($p < 0.0001$).

\subsection{Brain Age Gap}
Following much prior work \cite{Cole2018BrainMortality,Cole2017PredictingBiomarker,Peng2021AccurateNetworks,Lin2020OlderIllness,Smith2019EstimationImaging}, we use the high-dimensional biomarkers to develop a  scalar ``Brain Age Gap" (BAG), defined as the difference between a machine learning-predicted age and the subject's true age. In older adults, a large positive BAG can indicate more progressed neurodegeneration and is associated with various neurological conditions and other health outcomes \cite{Zhao2021LongitudinalLearning,Cole2018BrainMortality,Smith2019EstimationImaging,Lin2020OlderIllness}. Following \cite{J.2023TheAge}, we use a linear model fit to 1) R-NCE representations, 2) FS features, and 3) generic SSL representations from SimCLR on UKB subjects to calculate ``predicted ages". The differences between predicted ages and true age yields a BAG for each of the three feature types. We call these R-NCE-BAG (R-NCE Brain Age Gap), FS-BAG (FS Brain Age Gap), and SimCLR-BAG (SimCLR Brain Age Gap).

The brain age models are trained on an age and sex-balanced subset with 4000 subjects from UK Biobank. We first verify that the models give a good estimate of true age using the remaining 26,815 UK Biobank subjects for validation. The R-NCE-predicted brain age has a mean average error (MAE) of $4.05$ with Pearson correlation $r=0.724$. FS-predicted brain age has an MAE of $4.56$ years and $r=0.635$. This is in line with previous imaging-based brain age models using UK Biobank \cite{J.2023TheAge}. Finally, the SimCLR-predicted brain age has an MAE of $4.65$ years and $r=.611$. 

\begin{figure}[h]
\centering
\includegraphics[width=\linewidth]{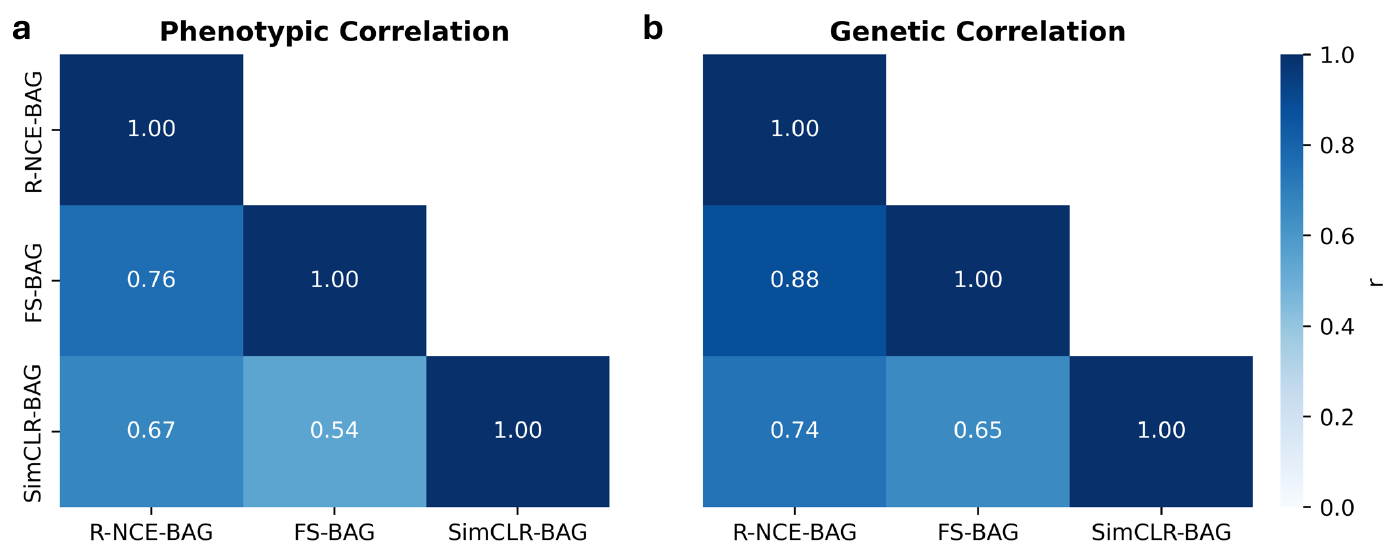}
\caption{Phenotypic and Genetic Correlation across BAGs} \label{Figure: BAG Correlation}
\end{figure}

After validating the goodness-of-fit of each BAG model, we measure the phenotypic correlation between the three BAG values, shown in Figure \ref{Figure: BAG Correlation}a. The two pairs with highest phenotypic correlation are between R-NCE/FS and R-NCE/SimCLR ($r = 0.76$, $r = 0.67$ respectively). FS and SimCLR BAGs have a slightly lower correlation ($r = 0.54$).

\begin{figure}[h]
\centering
\includegraphics[width=\linewidth]{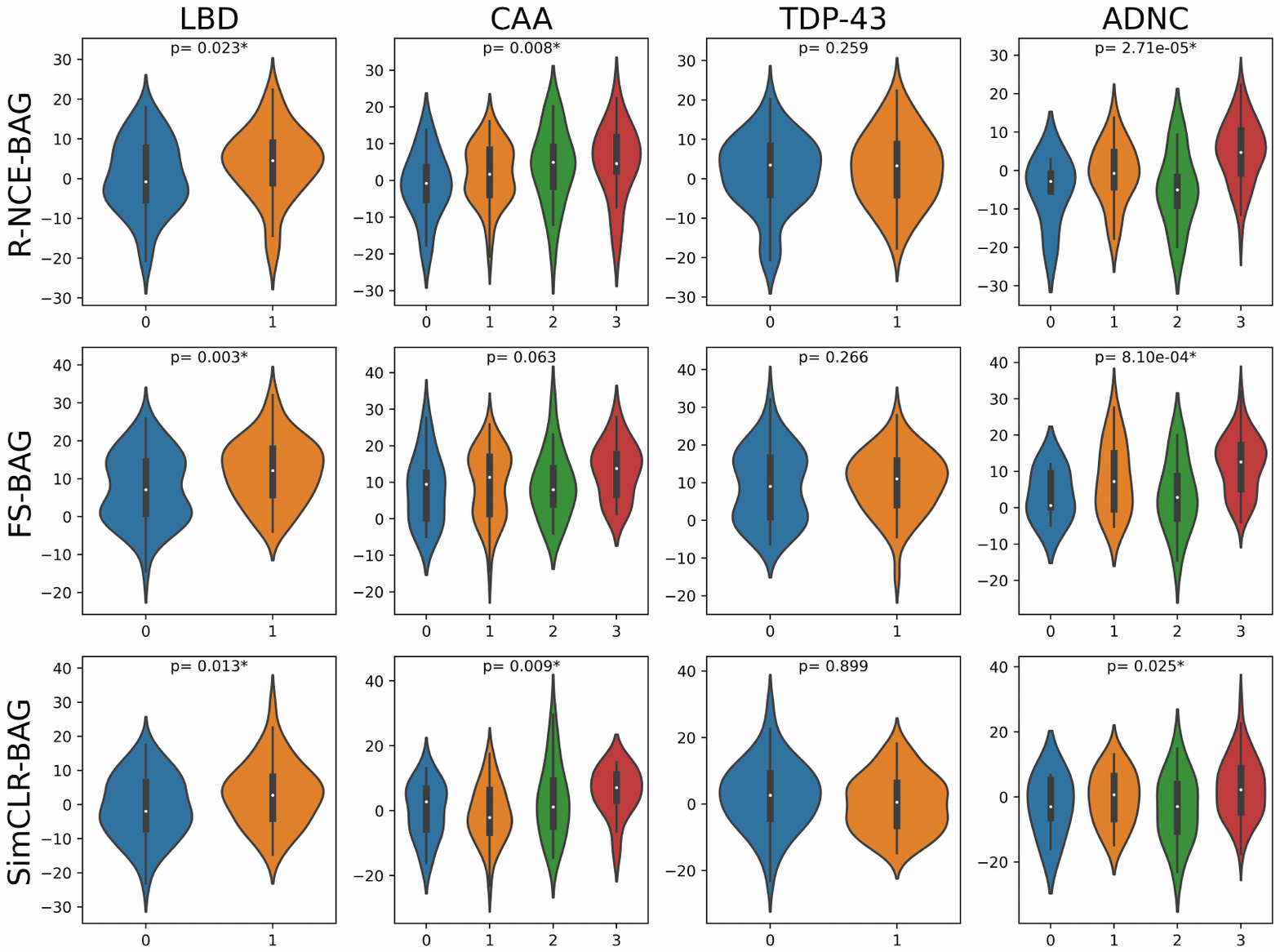}
\caption{Neuropathology outcomes (X-axis) and BAG scores. P-values are shown for each BAG + pathology combination, where * indicates statistical significance ($p < 0.05$ after false discovery rate correction)} \label{Figure: Neuropath}
\end{figure}

We first examine whether BAGs are associated with various AD and non-AD post-mortem neuropathological findings related to cognitive decline. 
Using the last T1-w MRI from subjects with post-mortem exams \cite{Tosun2024IdentifyingDisease}, we tested BAG associations with four neuropathologies. After FDR correction, all BAGs were associated with AD Neuropathological Changes (ADNC) and Lewy Body Disease (LBD), but none with transactive response DNA-binding protein 43 (TDP-43). Notably, R-NCE-BAG and SimCLR-BAG, but not FS-BAG, were associated with cerebral amyloid angiopathy (CAA) \ref{Figure: Neuropath}.

To identify genetic variants associated with accelerated neurodegeneration, we perform genome-wide association studies (GWAS) on each BAG  (see Section \ref{Methods: BAG GWAS} for details). Manhattan and QQ plots are shown in Supplementary Figure 1.
Using a p-value threshold of $5 \times 10^{-8}$, R-NCE-BAG, FS-BAG, and SimCLR-BAG are associated with 3, 6, and 6 genetic loci, respectively. We compute SNP-based heritability $h^2$ of each BAG using GCTA \cite{Yang2011GCTA:Analysis}. R-NCE-BAG ($h^2 = 0.42$), FS-BAG ($h^2 = 0.40$), and SimCLR-BAG ($h^2 = 0.37$) are all significantly heritable. We also compute the genetic correlation across the three BAG measures, shown in Figure \ref{Figure: BAG Correlation}b. The highest genetic correlation is between R-NCE-BAG and FS-BAG ($r = 0.88$), followed by R-NCE-BAG and SimCLR-BAG ($r = 0.74$) and FS-BAG and SimCLR-BAG ($r = 0.65$).

\begin{figure}[h]
\centering
\includegraphics[width=\linewidth]{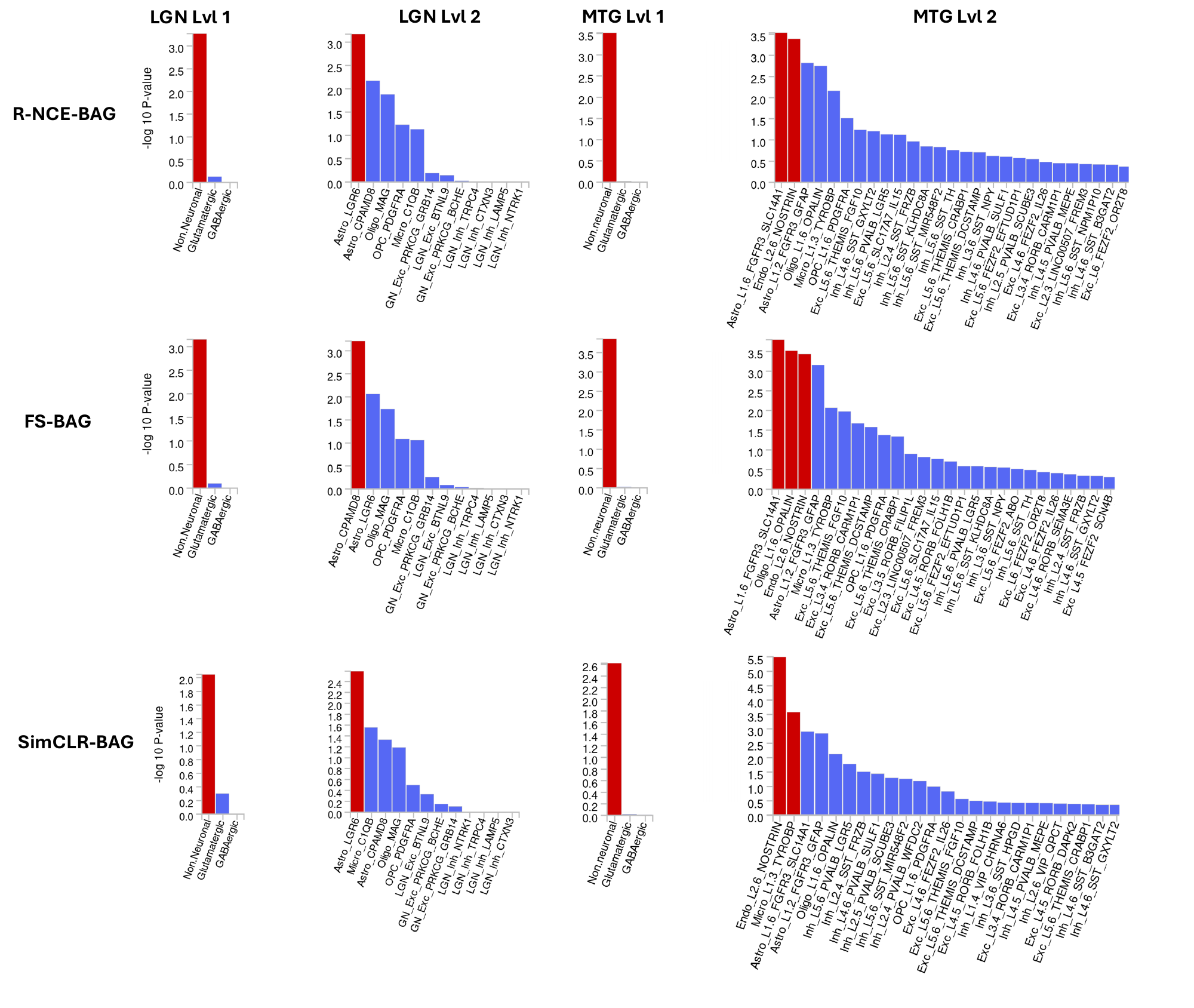}
\caption{Cell type enchrichment for R-NCE-BAG, FS-BAG, and SimCLR-BAG genes using scRNA-seq from lateral geniculate (LGN) and medial temporal gyrus (MTG). Significantly enriched cell types (using Bonferroni correction) are shown in red.} \label{Figure: Cell Types}
\end{figure}

We map associated variants to genes based on positional, expression quantitative trait loci (eQTL), and chromatin interaction information (See Section \ref{Methods: Gene Mapping}). R-NCE-BAG variants, FS-BAG variants, and SimCLR-BAG variants map to 153, 205, and 181 genes, respectively. We further map BAGs to different cell types using single-cell RNA sequencing (scRNA-seq) data \cite{Watanabe2019GeneticTraits}. We check for cell-type enrichment for BAGs using cell types defined by Allen Brain Institute cortical scRNA-seq in the lateral geniculate (LGN) and middle temporal gyrus (MTG). The results are shown in Figure \ref{Figure: Cell Types}. R-NCE, FS-BAG, and SimCLR-BAG genes are significantly enriched in non-neuronal cells. Both are enriched for LGN astrocytes. R-NCE-BAG genes are enriched in astrocyte and endothelial MTG cells, while FS-BAG is enriched in astrocyte, endothelial, and oligodendrodyte MTG cells. SimCLR-BAG genes are enriched in endothelial and microglia MTG cells.

\subsection{LDSC Regression}
\begin{figure}[h]
\centering
\includegraphics[width=\linewidth]{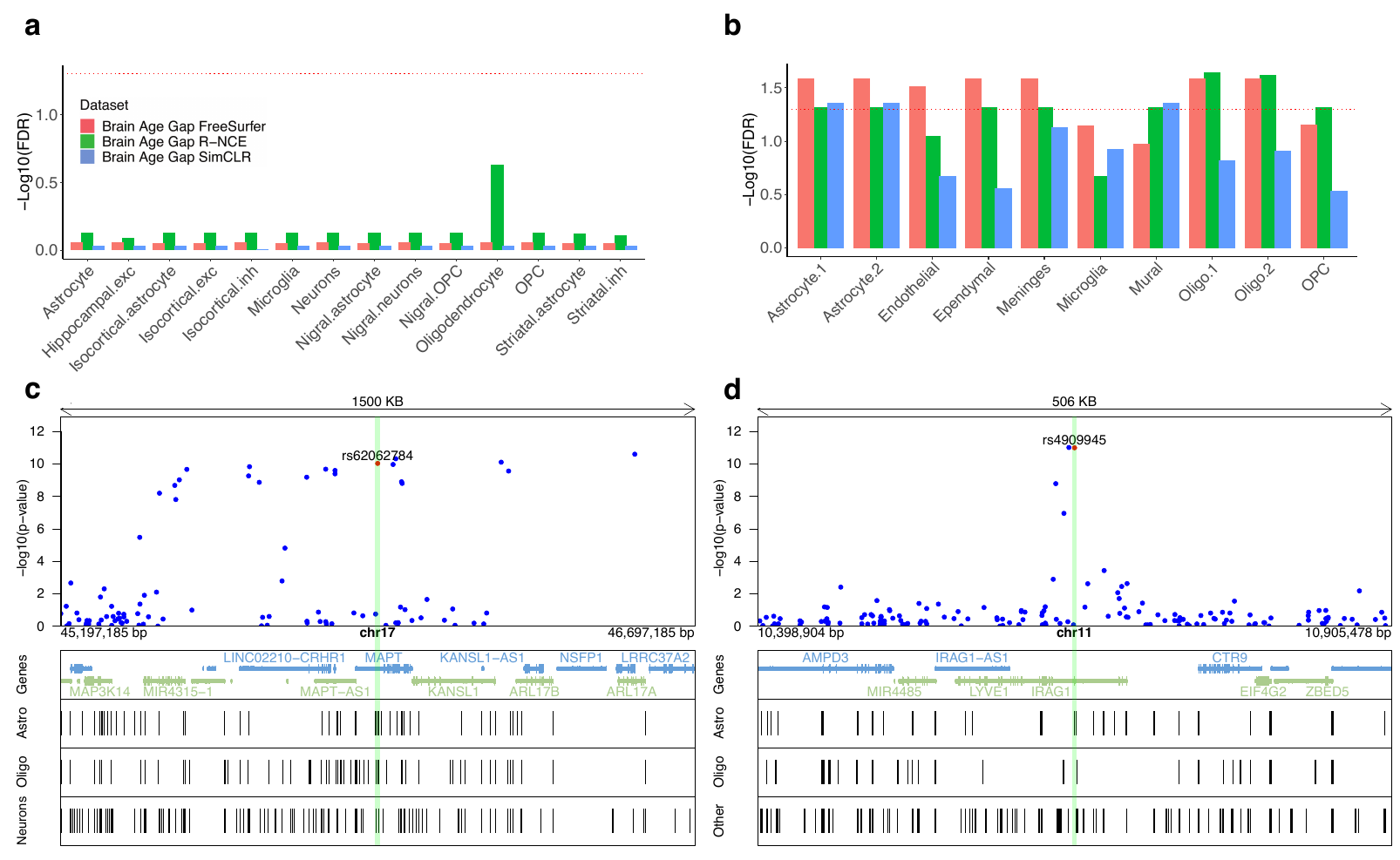}
\caption{Enrichment of BAG GWAS SNPs in Brain and Spinal Single-Nucleus Open Chromatin. a,b) Partitioned heritability of SNPs (stratified LDSC) by cell-type-specific open chromatin in brain a) and spinal cord b). The x-axis represents distinct cell types in each dataset. The y-axis is the -log10 transformation of the false-discovery-rated (FDR)-adjusted p-values of stratified LDSC significance. The red dotted line indicates the significance threshold FDR$<$0.05. Significance indicates that the set of open chromatin regions for a cell type explain significant heritability. c,d) Example BAG GWAS SNPs near genes linked to Alzheimer's Disease, MAPT and IRAG respectively. Top: Manhattan plots of significant SNPs (red dot, rsID above) with nearby SNPs in blue. Translucent green rectangles highlight the SNP falling within open chromatin. Middle: nearby genes. Bottom: open chromatin peaks of the highlighted cell types. SNP, single-nucleotide polymorphism; chr11,17, chromosome 11,17; bp, base-pair range of locus.} \label{Figure: LDSC}
\end{figure}

Functional annotations of genomic loci can provide valuable connections between variants of association from GWAS and their potential cellular effects \cite{Nasser2021Genome-wideGenes}. Gene expression patterns are driven by the activity of cis-regulatory elements (CREs) such as enhancers, which can be highly cell-type-specific \cite{Shlyueva2014TranscriptionalPredictions}. To further explore BAG variants in the context of epigenomic signatures, we use published datasets of cortex \cite{Corces2020Single-cellDiseases} and spinal \cite{Arokiaraj2024SpatialPredisposition} single-nucleus open chromatin datasets, based on the assay for transposase-accessible chromatin using sequencing (ATAC-seq).
First in brain, we apply stratified LDSC regression \cite{Finucane2015PartitioningStatistics} to identify the enrichment of R-NCE-BAG, FS-BAG, and SimCLR-BAG GWAS genetic variants in putative CREs of brain neuron and glial cell types (Figure \ref{Figure: LDSC}a).
In brain, we found a non-significant enrichment trend (R-NCE-BAG $>$ FS-BAG $>$ SimCLR-BAG). We repeated this in spinal glia, which have greater proportions of oligodendrocytes and astrocytes, and found significant enrichment for all cell types (Figure \ref{Figure: LDSC}b), with less enrichment of the SimCLR-BAG GWAS. Oligodendrocytes are most enriched in brain and spine, and for both the R-NCE-BAG and FS-BAG GWAS show greatest enrichment.

We also explore genome-wide-significant loci and their presence within cell-type-specific open chromatin. Within two genes, MAPT and IRAG1, we find genome-wide-significant SNPs that fall within open chromatin of astrocytes, oligodendrocytes, and oligodendrocyte precursor cells (OPCs) (MAPT, Figure \ref{Figure: LDSC}c), and within astrocytes (IRAG1, Figure \ref{Figure: LDSC}d).

\section{Discussion}\label{Discussion}

SSL has rapidly advanced as a powerful strategy for medical image representation, especially with the emergence of large foundation models \cite{Codella2024MedImageInsight:Imaging}. Yet, rigorous evaluation of SSL features remains underdeveloped, particularly in the context of AD. Most existing work benchmarks SSL representations using proxy tasks such as age prediction or generic disease classification \cite{Kim2023LearningImages,Zhao2021LongitudinalLearning}. However, clinically meaningful endpoints such as AD conversion prediction are often excluded. Furthermore, few studies explore cross-modal prediction tasks (e.g., estimating amyloid burden from MRI) that better reflect real-world clinical utility.


Another gap is the lack of comparison between SSL-derived and traditional morphometric features. Established volumetric and cortical thickness measures, such as those obtained from FS or ANTs \cite{Tustison2019LongitudinalStudy,Khadhraoui2024AutomatedFreeSurfer}, remain highly influential due to their interpretability and proven disease sensitivity. While supervised deep learning methods have been compared to FS features with mixed findings \cite{Liu2022GeneralizableMRIs,Arvidsson2024ComparingSymptoms}, similar benchmarking for SSL is absent. Such comparisons are essential before SSL approaches can be considered clinically viable.


To address these gaps, we conducted a comprehensive evaluation of SSL features compared to FS in the context of Alzheimer’s disease, including tasks related to age regression, disease classification, AD conversion, and amyloid status. In contrast to previous work \cite{Kim2023LearningImages,Zhao2021LongitudinalLearning,Kwak2023Self-SupervisedAmyloid-PET}, we trained all SSL methods on a large, independent pretraining dataset representative of the general population, then evaluated on a separate, disease-specific dataset. This design better reflects realistic deployment scenarios and mitigates overfitting, providing a fairer assessment of generalizability.


Across experiments, contrastive SSL methods (SimCLR and Barlow Twins) generally outperformed reconstruction-based methods (LSSL and Models Genesis). This likely reflects the advantage of augmentation-based objectives in capturing clinically relevant, invariant features, as opposed to voxel-level reconstruction. Nevertheless, FS features consistently outperformed all SSL approaches, particularly for the most clinically relevant endpoint of AD conversion prediction. These results indicate that current SSL methods, though powerful for representation learning, lack the disease specificity and biomarker sensitivity required for clinical translation.


To bridge the gap between FS and SSL, we developed R-NCE, a residual contrastive learning framework that explicitly incorporates auxiliary anatomical features, such as FS-derived measures, into SSL training. By aligning the learned representations with established biological features while preserving the benefits of contrastive learning in the residual space, R-NCE captures both domain-informed and augmentation-invariant information. This hybrid approach substantially improved downstream performance, surpassing both existing SSL methods and FS alone in multiple evaluation tasks, including 2- and 3-year AD conversion prediction. These findings suggest that integrating structured domain knowledge into SSL can produce representations that are both data-efficient and clinically meaningful.

We also used these features to create BAGs, a common marker of brain health \cite{Franke2019TenGained}. Unlike prior work comparing BAG models \cite{Peng2021AccurateNetworks} or modalities \cite{J.2023TheAge}, we held the modality (T1-w) and model (linear) constant, comparing BAGs derived from SimCLR, FS, and R-NCE features.
We found significant correlation between all BAGs, with phenotypic correlation ranging from 0.54-0.76 and genetic correlation ranging from 0.65-0.88, significantly higher than previous work studying cross-modality BAG correlation \cite{J.2023TheAge}. Despite high correlation, there remain some differences across R-NCE, FS, and SimCLR-BAGs. R-NCE-BAG and SimCLR-BAG are more sensitive to CAA neuropathology. R-NCE has the highest age-predictive accuracy and overall BAG heritability ($h^2 = 0.42$), indicating R-NCE-BAG may have the highest biological signal of the three. However, specific associated genetic loci varied across BAGs. Significant loci from all three BAGs were mapped to the MAPT gene, which encodes tau protein \cite{Goedert1989MultipleDisease}. Each BAG also mapped to additional unique loci. In scRNA-seq cell type matching, all BAGs showed enrichment in non-neuronal cell types and LGN astrocyte genes but slight difference in MTG cell type enrichment. 

We found enrichment of all BAG variants in the open chromatin of several distinct cell types, with the strongest trend (brain) and most significant results (spine) corresponding to oligodendrocytes and oligodendrocyte subtypes, respectively. FS and R-NCE variants showed consistently greater enrichment than SimCLR, providing additional evidence that FS and R-NCE have a stronger connection to the underlying cellular mechanisms of aging. A MAPT intronic variant is located within an open chromatin peak of several cell types, while a variant within IRAG1 is within an astrocyte-specific peak and found at the boundary of an intron and exon, functioning as a splice variant. MAPT's strong association with Alzheimer's disease is well documented. IRAG1 expression is abundant in astrocytes \cite{Zhang2022IdentificationProteomes} and the variant highlighted is associated with lacunar stroke \cite{Zhang2022IdentificationProteomes}, headache and migraine \cite{Meng2018A223773,Bjornsdottir2023RareAura}, and platelet levels \cite{Eicher2016Platelet-RelatedIndividuals}. While platelets and astrocytes are known to interact \cite{Li2022Astrocyte-DerivedPathway}, further study is needed to understand this variant's potentially causal role in brain aging and risk of stroke.

Our exploration of R-NCE has some limitations and opportunities for future exploration. 
First, we only studied AD using T1-w MRI.
Our R-NCE method should be applied to prediction of other neurological conditions including Parkinson's Disease or psychosis, which have known associations with structural imaging markers \cite{Zhu2024UsingHigh-risk,Pletcher2023CerebralDisease,Lin2024CorticalDisease}. Moreover, the method is generalizable to any imaging modality with associated image-derived features. Future work might apply R-NCE to diffusion MRI with associated features like anisotropy and mean diffusivity features or to PET imaging with regional uptake values as auxiliary features.



While promising, our study has several limitations. We focused exclusively on Alzheimer’s disease and T1-weighted MRI. Extending R-NCE to other neurological conditions—such as Parkinson’s disease or psychosis—and other modalities (e.g., diffusion MRI or PET) could test its generalizability. Additionally, while we evaluated SSL features in a fixed, linear setting, future work could explore fine-tuning R-NCE representations for disease-specific downstream tasks \cite{Kwak2023Self-SupervisedAmyloid-PET,Kaku2021IntermediateLearning}. Finally, the biological mechanisms driving differences between BAG measures warrant deeper investigation. The stronger enrichment patterns observed for FS and R-NCE variants suggest potential causal pathways that could be probed through molecular validation or multimodal integration.

Our findings highlight a critical message: generic SSL representations, though effective for broad image understanding, are not inherently sensitive biomarkers of neurodegeneration. By incorporating domain-informed auxiliary features, R-NCE closes this gap, producing imaging representations that are clinically predictive, biologically interpretable, and generalizable across datasets. This integration of SSL with structured anatomical knowledge offers a path forward toward clinically viable, biologically grounded self-supervised imaging biomarkers.

\section{Methods}\label{Methods}

\subsection{Patch-Based Model} \label{Methods: Patch-based Model}
We encode images using a 3D patch-based neural network model. This approach allows the model to learn 3D dependencies (rather than being constrained to 2-dimensional features). Using patches allows us to maintain high spatial resolution without needing to down-sample the image.

Image patches are encoded using a convolutional network encoder $f^l(X^l) \rightarrow H^l$ where $l$ represents the patch location and $X^l$ is the image patch \cite{Yu2023DrasCLR:Images,Sun2021ContextImages.}. The patch network takes variably sized input depending on the patch region. The network uses max-pooling and average-pooling to represent these differently sized patches with a fixed representation size. Although model architectures are constant across patch locations, separate patch model parameters are used for each patch location. The patch encoder \cite{Yu2023DrasCLR:Images} uses 13 convolution layers and the ELU activation \cite{Clevert2016FastELUs}. We also use batch normalization (synchronized across devices). Finally, we use a patch representation size of 64.

Patches are aggregated into an image-level representation using an aggregation model $f(\{H^l\}^{l \in L}) \rightarrow H^{\text{img}}$. We use a transformer encoder module \cite{Vaswani2017AttentionNeed} for $f$. Following ViT \cite{Dosovitskiy2020AnScale}, we prepend a learnable embedding token to the patch input and use its output representation as the image representation $H^{\text{img}}$. Additionally, learned positional encodings corresponding to patch location $l$ are added to each patch representation. The aggregation model uses a standard transformer encoder architecture, consisting of three transformer blocks, each with 2 attention heads. An image representation size of 64 is used.

\subsection{R-NCE}

Contrastive learning \cite{Oord2018RepresentationCoding} learns by discriminating between positive pairs (augmented versions of the same image) and negative pairs (different images). The Noise Contrastive Estimation (NCE) loss function is given by: 


\begin{equation} \label{Equation: nce_info}
    \mathcal{L}_{i,a}^{NCE} = - \log \frac{\exp \left[ \text{sim}( h_{i,a}, h_{i,a^c}) / \tau \right]}
    { \left[ \sum_{i',a'} \mathbb{I}\left[ i' \neq i \right] \exp \left[ \text{sim}( h_{i,a}, h_{i',a'} ) / \tau \right] \right]
    +\exp \left[ \text{sim}( h_{i,a}, h_{i,a^c} ) / \tau \right]}
\end{equation}

where $h_{i, a}$, $h_{i, a^c}$ are representations from two randomly augmented views of image $i$, encoded by a neural network $f(\cdot)$. Two augmentations are applied to each image in a given mini-batch. For a given image augmentation $a$ applied to an image, we denote $a^c$ as the other augmentation applied to the same image. 
A batch of images is used to draw negative samples $h_{i',a'}$. The $\text{sim}$ function is cosine similarity, $\tau$ is the temperature hyperparameter. The contrastive view of representation learning is closely related to information theory. It is straightforward to show $\text{I}(h_1, h_2) \geq -\mathcal{L}^{\text{NCE}} + \log(K)$, where $\text{I}(h_1,h_2)$ is the mutual information between $h_1$, $h_2$ \cite{Oord2018RepresentationCoding}. In other words, minimizing the NCE loss implicitly maximizes the mutual information between two random views. Since $\text{I}(h_1, h_2) = \text{H}(h_1,h_2) - \text{H}(h_1|h_2) - \text{H}(h_2|h_1)$, where $\text{H}(\cdot)$ denotes the entropy, maximizing the mutual information effectively maximizes the joint entropy of two variables.

We present a new framework for SSL with auxiliary information, called Residual-NCE (R-NCE).
Let  $a \in \mathbb{R}^{d_1}$ and $h \in \mathbb{R}^{d_2}$ denote the auxiliary information vector and the SSL representation vector respectively.  We use $A \in \mathbb{R}^{B \times d_1}$ and $H \in \mathbb{R}^{B \times d_2}$ to denote the matrix of all auxiliary information and SSL representations in a batch of size $B$. $A$ is a fixed matrix while $H$ is the output of the image encoder. 
Conceptually, we divide the representation into two subspaces, $\mathcal{A}$ and $\mathcal{A}^{\perp}$ (i.e., $h \in \mathcal{A} \oplus \mathcal{A}^{\perp}$), where $\mathcal{A}$ spans $a$ vectors and hence contains all the auxiliary information, and $\mathcal{A}^{\perp}$ contains the remaining information. 
Our goal is to make sure there is no information about $a$ outside of $\mathcal{A}$ and that the information content of $\mathcal{A}^{\perp}$ is maximized.
To do this, we define two residuals:
\begin{eqnarray} \label{Equation: Double Residuals}
 R = A - P_{H}[A] \nonumber\\
 R' = H - P_{A}[H],
\end{eqnarray}
where $P_{X}[Y] = XX^{\dagger}Y$ is the projection matrix on the subspace that spans $X$ applied on matrix $Y$. Note that the residuals are mirrors of each other. 
By minimizing the norm of $R$, we ensure all the information in $A$ is already in a subspace of $H$. By applying the NCE loss on $R'$, we ensure the joint entropy in the remaining space (after removing all side information) contains maximal information.
The total loss can be written as:
\begin{equation}
  \mathcal{L}^{\text{R-NCE}} = \| R \|_{F}^2 + \mathcal{L}^{NCE}(R')
\end{equation}
where $\mathcal{L}^{NCE}(R')$ is the loss from (\ref{Equation: nce_info}), applied on the residualized embedding space.

\subsection{Image Preprocessing}
Before the multi-level SSL training process, we first discuss the image preprocessing and patchification pipeline. Images are preprocessed with motion correction, intensity normalization, and skull-stripping using the FS version 7.1.1 recon-all pipeline \cite{Fischl2012FreeSurfer}. Next, we register images to the MNI-152 template using a similarity (9 degrees of freedom) transformation with advanced normalization tools (ANTs) \cite{Avants2011ARegistration}. Then, 18 overlapping patches from each image are extracted according to regions of the brain, namely, the left and right medial temporal, lateral temporal, parietal, occipital, subcortical, cingulate, medial frontal, anterior frontal, and posterior frontal areas. The patch sizes for all 18 regions are shown in Supplementary Table 1.

Cortical thickness and subcortical volume features from images are also obtained from FS recon-all. The features are harmonized across datasets and scanners using ComBat \cite{Raamana2020ConqueringConfounds,Fortin2018HarmonizationSites}.

For SSL training we apply the following random augmentations at the patch level: rotation, crop, noise, gaussian blur, bias field, flip, spike, contrast adjustment, and mapping of grey matter/white matter/CSF labels to random Gaussian distributions \cite{Billot2021SynthSeg:Resolution}. These transforms are applied using Monai \cite{Cardoso2021MONAI:Healthcare} and TorchIO \cite{Perez-Garcia2021TorchIO:Learning} software packages.

\subsection{Datasets}
For all SSL experiments, we use separate datasets for training and evaluation. While many SSL studies use the same dataset for SSL training and evaluation, we want to thoroughly evaluate the generalizability of our methods to external images with different population characteristics and scanner parameters. We use the UK Biobank (UKB) \cite{Sudlow2015UKAge} and Oasis-3 \cite{LaMontagne2019OASIS-3:Disease} datasets for SSL training. ADNI is used for evaluation. The SSL dataset (UKB + Oasis-3) contains 32,132 unique subjects with, at baseline, 80.8\% unknown diagnosis (UKB), 13.0\% cognitively normal (Oasis-3), and 6.2\% early-stage Alzheimer's disease (eAD) (Oasis-3). Although UKB subjects do not have diagnosis labels, the study participants are largely healthy even compared to the general UK population \cite{Fry2017ComparisonPopulation}.

The evaluation dataset (ADNI) contains 2,431 subjects with, at baseline, 36.8\% CN, 46.0\% MCI, 17.2\% AD. SSL dataset subjects are 47.2\% male with an average age of 64.3 (std=7.9). Evaluation dataset subjects are 51.5\% male with an average age of 73.0 (std=7.5). Additional dataset demographic information is shown in Table \ref{Table: Demographics}.

\begin{table}[!h]
\centering
\caption{Summary of Datasets Used. We use the SSL Dataset (UKB + OASIS-3) for self-supervised learning (SSL) training. The evaluation dataset (ADNI) is used for all evaluations, where a linear head is trained on the frozen pre-trained encoder.}\label{Table: Demographics}
\begin{tabular}{cccccc}
\hline
\textbf{}                                                                      & \# Subjects & \, \# Images \, & Age        & \, Gender \,                                                      & Diagnosis                                                                          \\ \hline
\textbf{UKB}                                                                   & 30,815      & 33,080    & 64.1 (7.7) & \begin{tabular}[c]{@{}c@{}}47.3\% M\\ 52.7\% F\end{tabular} & 100\% Unknown                                                                      \\ \hline
\textbf{OASIS-3}                                                               & 1,316       & 2,681     & 70.1 (9.3) & \begin{tabular}[c]{@{}c@{}}44.5\% M\\ 55.5\% F\end{tabular} & \begin{tabular}[c]{@{}c@{}}56.2\% CN\\ 43.8\% Dementia\end{tabular}                \\ \hline
\textbf{\begin{tabular}[c]{@{}c@{}}SSL Dataset\\ (UKB + OASIS-3)\end{tabular}} & 32,131      & 35,761    & 64.3 (7.9) & \begin{tabular}[c]{@{}c@{}}47.2\% M\\ 52.8\% F\end{tabular} & \begin{tabular}[c]{@{}c@{}}95.9\% Unknown\\ 2.3\% CN\\ 1.8\% Dementia\end{tabular} \\ \hline
\textbf{\begin{tabular}[c]{@{}c@{}}Evaluation Dataset\\ (ADNI)\end{tabular}}   & 2,431       & 10,620    & 73.0 (7.4) & \begin{tabular}[c]{@{}c@{}}51.5\% M\\ 48.5\% F\end{tabular} & \begin{tabular}[c]{@{}c@{}}36.8\% CN\\ 46.0\% MCI\\ 17.2\% AD\end{tabular}         \\ \hline
\end{tabular}
\end{table}

\subsection{Model Training}
\begin{figure}[h]
\includegraphics[width=\textwidth]{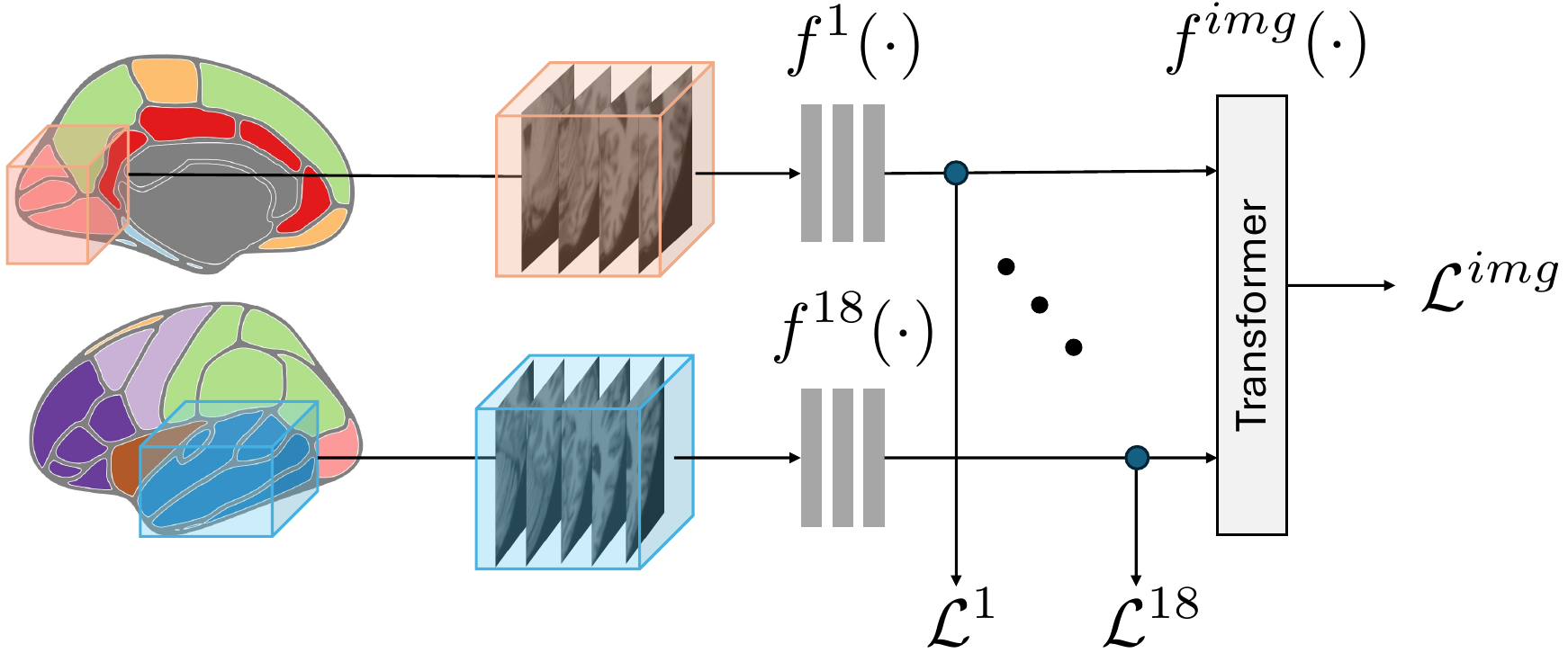}
\caption{Encoder architecture. 3D patches are extracted from 9 regions in each hemisphere (8 cortical, 1 subcortical). Patches are fed through separate encoders $f^{1}$,$f^{2}$,...,$f^{18}$, then patch representations are aggregated with a transformer encoder $f^{img}$. The loss $\mathcal{L}$ is applied at both patch ($\mathcal{L}^{1}$,$\mathcal{L}^{2}$,...,$\mathcal{L}^{18}$) and image ($\mathcal{L}^{img}$) levels. }\label{Figure: Multilevel Loss}
\end{figure}

We perform SSL training as a multi-stage process at two levels. First, we train at the patch level; next we use pre-trained patch encoders to train at the aggregate image level. A simplified depiction of the overall training process is shown in Figure \ref{Figure: Multilevel Loss}.

First, each patch model is trained separately. The patch models (one corresponding to each patch location) have separate parameters, allowing each model to be trained in parallel. Patches are first extracted from the image corresponding to location $l$. Each patch is randomly augmented twice to produce a positive pair $(X_{i,a}^l,X_{i,a^c}^l)$ which is passed through the patch model (along with negative samples $X_{i',a'}^l$) to obtain $H^l$. 

Importantly, the auxiliary features $A$ used in $\mathcal{L}^{aux}$ at the patch representation level $H^l$ include only the imaging features corresponding to the brain region covered by the specific patch. For example, the parietal patch representations should be predictive of postcentral, supramarginal, superiorparietal, inferiorparietal, and precuneus thickness.

After all patch models have been trained, the aggregation model is trained using representations $\{H^l\}^{l \in L}$ from the frozen patch models. To augment aggregation model input, we randomly split patch representations into two subsets. Specifically, in each training batch, each patch location is randomly assigned to subset 1 $S1$ or subset 2 $S2$ with $50\% $ probability. $\{H^l\}^{l \in S1}$ and $\{H^l\}^{l \in S2}$ for each image form positive samples for $\mathcal{L}^{NCE}$; negative samples are subsets of patch representations within the batch from different images. The subsets of patch representations are fed into the ViT aggregation model to yield a single image representation for each subset. The ViT aggregation model naturally allows for an arbitrary number of patch representations as input, so each patch subset can be encoded as separate image representations. 

As in patch model training, the contrastive loss $\mathcal{L}^{\text{NCE}}$ is applied on the positive and negative sample image representations. Only auxiliary features $A$ covered by the corresponding patch subsets are used in $\mathcal{L}^{\text{R-NCE}}$ (i.e. as auxiliary features $A$) for each image representation. 

\subsection{Baseline SSL}\label{Methods: Baseline}
For all representation quality evaluation we evaluate against different SSL methods and against using FS cortical thickness and subcortical volume measurements alone. The first method we evaluate against is SimCLR \cite{Chen2020ARepresentations}, a contrastive method which uses the standard temperature-scaled NCE loss shown in (\ref{Equation: nce_info}). Our next baseline method is Barlow Twins \cite{Zbontar2021BarlowReduction}, a ``dimension-contrastive'' \cite{Garrido2022OnLearning} method which minimizes covariance across embedding dimensions (rather than using negative samples). We also evaluate against LSSL \cite{Zhao2021LongitudinalLearning}, a autoencoder-based SSL method where intra-subject longitudinal trajectories are aligned in the latent space. Finally, we use Models Genesis \cite{Zhou2021ModelsGenesis}, a foundational method for medical imaging SSL where images are cropped and deformed, then the network is trained to generate the original sub-volume.

The aforementioned four baselines represent ``generic" SSL approaches. In other words, while some of the methods are tailored specifically for medical images and some are adapted from the natural image setting, all of these baselines do not use any auxiliary side information. 

Besides other SSL methods, we use FS features as a baseline method for comparison. We use the standard 104 cortical thickness and subroctical volume features from the FS version 7.1.1 recon-all pipeline, harmonized using ComBat \cite{Raamana2020ConqueringConfounds,Fortin2018HarmonizationSites}.

For all SSL methods besides reconstruction-based approaches (namely R-NCE, SimCLR, Barlow Twins), we use identical neural network architectures (see \ref{Methods: Patch-based Model}). Specifically, we keep the patch and aggregation models constant across the methods and apply the loss at both the patch and image levels. The sequential training scheme (training patch models separately, then aggregation models) is also kept constant for these methods.

For reconstruction-based methods (LSSL, Models Genesis), we use whole-image models and apply the corresponding SSL loss functions at a single level. We choose this approach because reconstruction-based methods are not easily adaptable to the multi-level loss. Specifically, the decoding of a high-resolution whole-image input (i.e. the input to all patch encoders) is not computationally feasible. Therefore, for LSSL and Models Genesis we used a downsampled and cropped $96 \times 96 \times 96$ voxel image as input to the autoencoder and U-Net respectively. For architecture and input details for all methods, see Supplementary Table 2.
Finally, the training hyperparameters for all methods are shown in Supplementary Table 3.

\subsection{Biomarker Sensitivity Comparison}
For all experiments in Section \ref{Results: Sensitivity}, we obtain AUROC or $R^2$ values from 1000 bootstrap samples, fitting a linear head on 2,331 patients and leaving out 100 patients for evaluation in each sample. To assess the statistical significance of differences in classifier performance, we conducted a permutation test on the AUROC or $R^2$ values. Given two models, we first compute the observed mean AUROC or $R^2$ difference, then generate a null distribution by randomly permuting AUROC or $R^2$ scores between models and recalculating the difference 10,000 times. The p-value is the proportion of permuted differences with an absolute value greater than or equal to the observed difference.


\subsection{Brain Age Modeling}
SSL features (from R-NCE and SimCLR) and FS cortical thickness and subcortical volume features are used to fit three separate linear regression models to predict age. Following prior work \cite{J.2023TheAge}, we identify 4000 UK Biobank subjects comprising 8 buckets corresponding to age (44-53, 54-63, 64-73, 74-83) and sex (M, F) for fitting the brain age model. For subjects with longitudinal images, only their initial images are used for BAG analysis. We debias the age prediction model by orthogonalizing the brain age gap with respect to true age \cite{Smith2019EstimationImaging}.
\subsection{GWAS} \label{Methods: BAG GWAS}
We perform similar GWAS quality control to prior work \cite{J.2023TheAge,Marees2018AAnalysis}. We remove subjects with discrepancies between self-identified and genetic sex. We retain only autosomal SNPs. We use the KING software to remove related (2nd degree) individuals \cite{Manichaikul2010RobustStudies}. Next, we filter SNPs with missingness greater than 3\%, SNPs with minor allele frequency less than 1\%, and duplicated SNPs. We then remove variants that fail Hardy-Weinberg Equilibrium test at $1 \times 10^{-10}$. We also remove individuals with SNP missingness greater than 3\%.

For GWAS analysis, we fit linear models using Plink2 \cite{Purcell2007PLINK:Analyses}. We include the following covariates: Age, Age$^2$, Sex, Age $\times$ Sex, Age$^2$ $\times$ Sex, and Total Intracranial Volume. We also include the first 20 genetic principal components to account for population stratification.

\subsection{Gene Mapping} \label{Methods: Gene Mapping} Genetic loci and gene mapping using GWAS summary statistics was performed using FUMA \cite{Watanabe2017FunctionalFUMA}. SNPs are mapped to genes using 1) positional mapping (with a maximum distance of 10kb) 2) eQTL Mapping  and 3) 3D chromatin interaction mapping. scRNA-seq cell type enrichment is also performed using FUMA \cite{Watanabe2019GeneticTraits} with data from Allen Human Brain cortical samples \cite{Hodge2018ConservedCortex}. 

\subsection{LDSC Regression} \label{Methods: LDSC Regression} 
We munged R-NCE BAG, FS BAG, and SimCLR BAG GWAS using the following function in LDSC \cite{Bulik-Sullivan2015LDStudies}:
\begin{verbatim} 
./munge_sumstats.py --signed-sumstats BETA,0 
--out ../munged/BrainAgeGap-Batmanghelich_2024 --N 25825 --a1 
A1 --a2 REF --snp ID --sumstats GWAS_SUMMARY_STATISTICS --p P
\end{verbatim}

We utilize LD score annotations of human isocortex, striatum, hippocampus, and substantia nigra single cell chromatin accessibility measurements \cite{Corces2020Single-cellDiseases}  from \cite{Srinivasan2021Addiction-associatedNeurobiology}. We also utilize LD score annotation of mouse-human conserved dorsal horn single cell chromatin accessibility measurements from \cite{Arokiaraj2024SpatialPredisposition}. Following the LDSC Regression pipeline described in \cite{Srinivasan2021Addiction-associatedNeurobiology}, we estimate the conditional heritability enrichment of R-NCE-BAG, FS-BAG, and SimCLR-BAG genetic variants across open chromatin regions of cell populations measured in 1) \cite{Corces2020Single-cellDiseases}: astrocytes, hippocampal excitatory neurons, isocortical astrocytes, isocortical excitatory neurons, isocortical inhibitory neurons, microglia, neurons, nigral astrocytes, nigral neurons, nigral oligodendrocyte precursor cells (OPC), striatal astrocytes, and striatal inhibitory neurons and 2) \cite{Arokiaraj2024SpatialPredisposition}: astrocytes (Astrocyte.1, Astrocyte.2), endothelial cells, epyndymal cells, meninges, microglia, mural cells, oligodendrocytes (Oligodendrocyte.1, Oligodendrocyte.2), and OPC. We adjust for multiple hypothesis testing using the false discovery rate (FDR) on p-values of the LDSC regression coefficients ($\alpha = 0.05$) across all 3 BAG GWAS and sets of open chromatin regions for a single dataset.

To identify particular loci with overlap of cell-type-specific open chromatin, we first limit to SNPs with unadjusted significance of p$<$0.05 and sorted by ascending p-value. We then intersect these SNPs with peaks from cell-type-specific chromatin using the command \%over\% from the package GRanges in R.
We visualize SNPs overlapping open chromatin regions and genes (Figure \ref{Figure: LDSC}c,d) with the R package plotgardener \cite{Kramer2022Plotgardener:R}.

\backmatter

\clearpage
\bibliography{references}

\end{document}


\title{Supplementary Information for:
A Cautionary Tale of Self-Supervised Learning for Imaging Biomarkers: Alzheimer's Disease Case Study}

\author{Maxwell Reynolds et al.}

\maketitle
\setcounter{section}{1}

\section*{Supplementary Information}

\subsection{Data and Model Specifications}

\begin{table}[!h]
\centering
\caption{Patch sizes for the 18 locations used in this study. These overlapping locations are defined using McGill CerebrA atlas labels.}\label{Supplementary Table: Patch Sizes}
\begin{tabular}{ccc}
\hline
                          & \textbf{\: \: \: \: \: \: \: \:Left \: \: \: \: \: \: \: \:}  & \textbf{\: \: \: \: \: \: \: \:Right \: \: \: \: \: \: \: \:} \\ \hline
\textbf{MTL}              & 57 x 107 x 64  & 57 x 107 x 64  \\ \hline
\textbf{LTL}              & 70 x 115 x 90  & 70 x 115 x 90  \\ \hline
\textbf{Occipital}        & 79 x 87 x 82   & 80 x 87 x 82   \\ \hline
\textbf{Parietal}         & 86 x 108 x 94  & 87 x 108 x 94  \\ \hline
\textbf{Subcortical}      & 63 x 90 x 75   & 63 x 90 x 75   \\ \hline
\textbf{Cingulate}        & 40 x 122 x 82  & 41 x 122 x 82  \\ \hline
\textbf{Medialfrontal}    & 50 x 138 x 128 & 50 x 138 x 128 \\ \hline
\textbf{Posteriorfrontal} & 78 x 98 x 99   & 78 x 98 x 99   \\ \hline
\textbf{Anteriorfrontal}  & 74 x 84 x 104  & 74 x 84 x 104  \\ \hline
\end{tabular}
\end{table}

\begin{table}[!h]
\centering
\caption{Description for models used in the study. Besides generative methods (LSSL, Models Genesis), a consistent multi-level modeling scheme is used. The multi-level modeling is not well-defined for generative methods, as it would require decoding the entire high-resolution image. The generative models instead take as input a $96 \times 96 \times 96$ image of the entire brain.}\label{Supplementary Table: Models}
\begin{tabular}{ccc}
\hline
\textbf{SSL Method(s)}                                                             &\: \: \: \: \: \: \: \textbf{Input}         \: \: \: \: \: \: \:                                         & \: \: \: \: \: \: \: \: \: \: \:  \: \: \textbf{Model} \: \: \: \: \: \: \: \: \: \:                                                                                                                     \\ \hline
\begin{tabular}[c]{@{}c@{}}SimCLR, \\ Barlow Twins, \\ R-NCE\end{tabular} & \begin{tabular}[c]{@{}c@{}}Regional \\ Image Patch\end{tabular} & \begin{tabular}[c]{@{}c@{}}12-layer ConvNet, \\ BatchNorm, ELU\end{tabular}                                                        \\ \hline
\begin{tabular}[c]{@{}c@{}}SimCLR, \\ Barlow Twins, \\ R-NCE\end{tabular} & Patch Embeddings                                                & \begin{tabular}[c]{@{}c@{}}ViT Encoder. 3 transformer blocks \\ (each with 2 attention heads)\end{tabular}                         \\ \hline
LSSL                                                                               & \begin{tabular}[c]{@{}c@{}}96 x 96 x 96 \\ Image\end{tabular}   & \begin{tabular}[c]{@{}c@{}}Autoencoder \\ (w/ 4-layer ConvNet encoder), \\ BatchNorm, ReLU\end{tabular}                            \\ \hline
Models Genesis                                                                     & \begin{tabular}[c]{@{}c@{}}96 x 96 x 96 \\ Image\end{tabular}   & \begin{tabular}[c]{@{}c@{}}U-Net \\ (w/ 8-layer ConvNet encoder), \\ BatchNorm, ReLU, \\ (Original Models Genesis Model)\end{tabular} \\ \hline

\end{tabular}
\end{table}

\begin{table}[!h]
    \centering
    \caption{Hyperparameters for each method. A grid search is used to determine learning rate (lr) for each method. Temperature ($\tau$) was fixed at 0.5 \cite{Chen2020ARepresentations}. Batch size was maximized for all methods.}
    \begin{tabular}{ccc}
        \hline
        \textbf{Method} & \: \: \: \: \: \textbf{Level}\: \: \: \: \:  & \: \: \: \: \: \: \: \: \: \: \: \: \textbf{Hyperparameters}\: \: \: \: \: \: \: \: \: \: \: \:  \\
        \hline
        SimCLR & Patch & lr = 3.0e-04, $\tau$ = 0.5, batch size = 128 \\
        \hline
        SimCLR & Aggregate & lr = 3.0e-05, $\tau$ = 0.5, batch\_size = 256 \\
        \hline
        Barlow Twins & Patch & lr = 3.0e-04, $\lambda$ = 5.0e-03, batch size = 128 \\
        \hline
        Barlow Twins & Aggregate & lr = 3.0e-05, $\lambda$ = 5.0e-03, batch size = 256 \\
        \hline
        LSSL & Image & lr = 0.001, $\lambda$ = 1, batch size = 16 \\
        \hline
        Models Genesis & Image & lr = 1, batch size = 6 \\
        \hline
        R-NCE & Patch & lr = 3.0e-05, $\tau$ = 0.5, batch size = 128 \\
        \hline
        R-NCE & Aggregate & lr = 3.0e-05, $\tau$ = 0.5, batch size = 256 \\
        \hline
    \end{tabular}
    \label{Supplementary Table: Hyperparameters}
\end{table}

\clearpage
\subsection{GWAS Results}
\begin{figure}[!h]
    \centering
    \includegraphics[width=\textwidth]{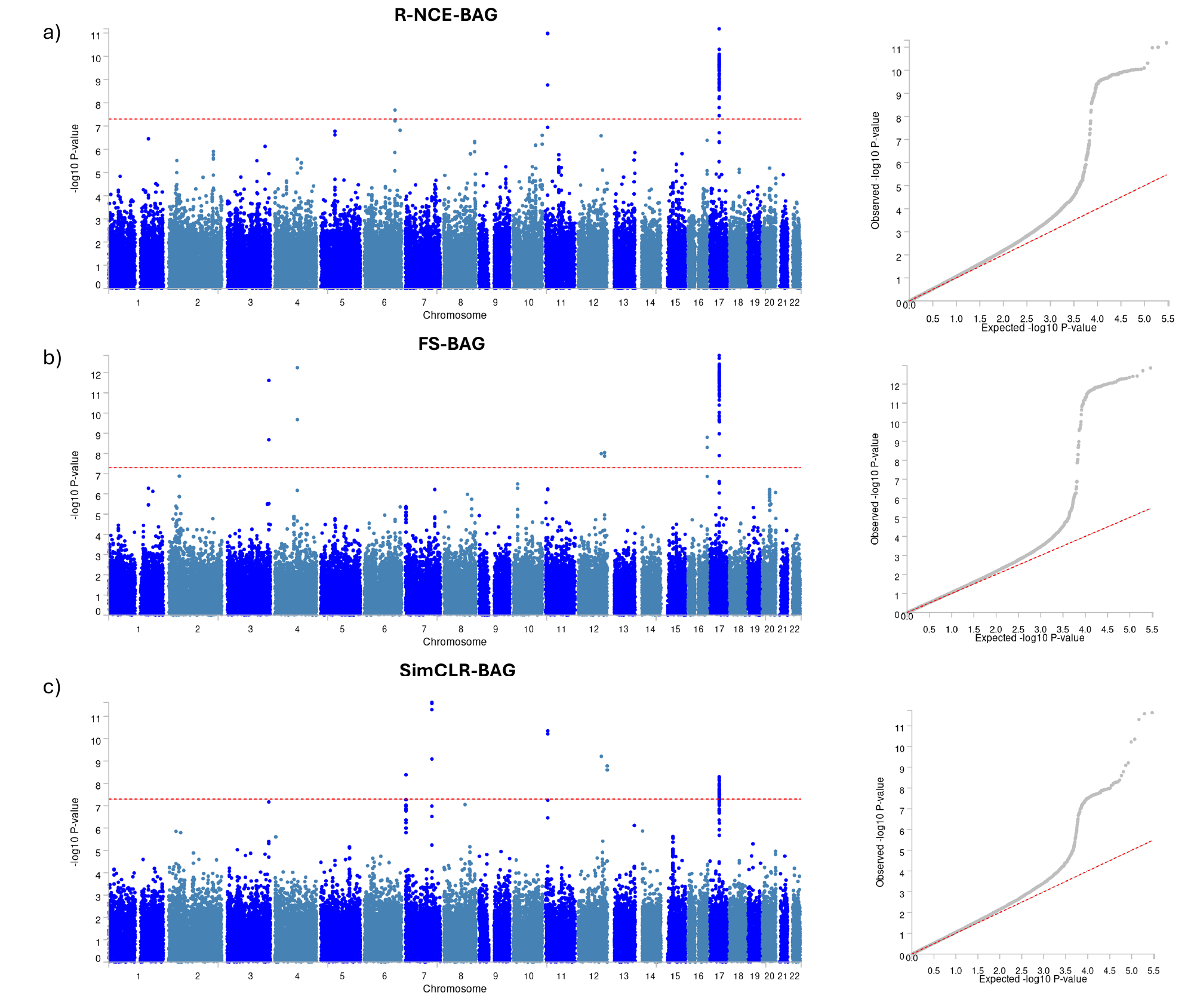}
    \caption{Manhattan Plots and QQ plots for a) R-NCE-BAG and b) FS-BAG, and c) SimCLR-BAG}
    \label{Figure: Manhattan and QQ}
\end{figure}

\bibliography{references}